\documentclass[a4paper]{article}

\usepackage[english]{babel}
\usepackage[utf8x]{inputenc}
\usepackage[T1]{fontenc}
\usepackage{authblk}

\usepackage[a4paper,top=3cm,bottom=2cm,left=3cm,right=3cm,marginparwidth=1.75cm]{geometry}

\usepackage{booktabs}       %
\usepackage{amsfonts}       %
\usepackage{nicefrac}       %
\usepackage{microtype}      %
\usepackage{diagbox} %
\usepackage{dsfont}
\usepackage{bm} %

\usepackage{amsmath}
\usepackage{graphicx}
\usepackage{amsfonts}
\usepackage{subcaption}
\usepackage{stmaryrd}
\usepackage{color, colortbl}
\usepackage{breakcites}
\usepackage{pdfpages}
\usepackage{rotating}
\usepackage{dirtree} %
\usepackage[colorinlistoftodos]{todonotes}
\usepackage[colorlinks=true, allcolors=blue]{hyperref}
\usepackage{multirow}
\usepackage{soul}
\newtheorem{mydef}{Definition}
\newtheorem{recommendation}{Recommendation}
\newtheorem{constraint}{Constraint}
\newtheorem{relaxation}{Relaxation}
\newtheorem{desideratum}{Desideratum}

\usetikzlibrary{arrows}
\usepackage{color, colortbl} %
\usepackage{multicol}
\usepackage{amssymb}
\usepackage{enumitem}
\definecolor{Gray}{gray}{0.9}
\definecolor{LightCyan}{rgb}{0.88,1,1}  %

\usepackage{array}

\newcommand{\vin}[1]{\textcolor{black}{#1}}
\newcommand{\tim}[1]{\textcolor{black}{#1}}
\newcommand{\nat}[1]{\textcolor{black}{#1}}

\newcolumntype{P}[1]{>{\centering\arraybackslash}p{#1}}

\title{Continual Learning for Robotics: Definition, Framework, Learning Strategies, Opportunities and Challenges} %

\date{\vspace{-5ex}}

\author[*,1,2]{Timoth\'ee Lesort}
\author[*,3]{Vincenzo Lomonaco}
\author[2]{Andrei Stoian}
\author[3]{Davide Maltoni}
\author[1]{David Filliat}
\author[*,1]{Natalia D\'iaz-Rodr\'iguez}%
\affil[1]{Flowers Team (ENSTA Paris, Institut Polytechnique de Paris \& INRIA).}%
\affil[2]{Thales, Theresis Laboratory.}%
\affil[3]{Department of Computer Science and Engineering - University of Bologna}%
\affil[*]{Equal contribution.}%

\providecommand{\Keywords}[1]{\textbf{\textit{Keywords: }} #1}
\begin{document}
\maketitle

\begin{abstract}

Continual learning (CL) is a particular machine learning paradigm where the data distribution and learning objective change through time, or where all the training data and objective criteria are never available at once.
The evolution of the learning process is modeled by a sequence of learning experiences where the goal is to be able to learn new skills all along the sequence without forgetting what has been previously learned. CL can be seen as an online learning where knowledge fusion needs to take place in order to learn from streams of data presented sequentially in time.
Continual learning also aims at the same time at optimizing the memory, the computation power and the speed during the learning process.

An important challenge for machine learning is not necessarily finding solutions that work in the real world but rather finding stable algorithms that can \emph{learn} in real world.
Hence, the ideal approach would be tackling the real world in a embodied platform: an autonomous agent.
Continual learning would then be effective in an autonomous agent or robot, which would learn autonomously through time about the external world, and incrementally develop a set of complex skills and knowledge.

Robotic agents have to learn to adapt and interact with their environment using a continuous stream of observations.
Some recent approaches aim at tackling continual learning for robotics, but most recent papers on continual learning only experiment approaches in simulation or with static datasets. Unfortunately, the evaluation of those algorithms does not provide insights on whether their solutions may help continual learning in the context of robotics. 
This paper aims at reviewing the existing state of the art of continual learning, summarizing existing benchmarks and metrics, and proposing a framework for presenting and evaluating both robotics and non robotics approaches in a way that makes transfer between both fields easier. We put light on continual learning in the context of robotics to create connections between fields and normalize approaches.

\end{abstract}



\section{Introduction}
Machine learning (ML) approaches generally learn from a stream of data randomly sampled from a stationary data distribution. This is often a \textit{sine qua non} condition to learn efficiently. However, in the real world, this setting is rather uncommon. \textit{Continual Learning} (CL) \cite{ring94} gathers together work and approaches that tackle the problem of learning when the data distribution changes over time, %
and where knowledge fusion over never-ending streams of data needs to be accounted for. Consequently, CL is the paradigm to deal with catastrophic forgetting \cite{Mccloskey89,French99}.

For convenience, we can empirically split the data stream into several subsections temporally bounded we call tasks.
We can then observe what we learn or forget when learning a new task.
Even if there is no mandatory constraint on a task, a task often refers to a particular period of time where the data distribution may (but not necessarily) be stationary, and the objective function constant. Tasks can be disjoint or related to each other, in terms of learning objectives, and depending on the setting.

One solution to Continual Learning would be saving all data, shuffle it, and come back to a traditional machine learning setting. Unfortunately, in this case, this is not always possible nor optimal. Here are several examples of settings where continual learning is necessary:

\begin{itemize}
    \item You have a trained model, you want to update it with new data but the original training data was discarded or you do not have the right to access it any longer.
    \item You want to train a model on a sequence of tasks but you can not store all your data or you do not have the computational power to retrain the model from all data (e.g., in an embedded platform).
    \item You want an agent to learn multiple policies but you do not know when the learning objective changes nor how.
    \item You want to learn from a continuous stream of data that may change through time but you do not know how and when.
\end{itemize}

In order to handle such settings, representations should be learned in an online manner \cite{Li17}. As data gets discarded and has a limited lifetime, the ability to forget what is not important and retain what matters for the future are the main issues that continual learning targets and focuses on.

From a robotics point of view, CL is the machine learning answer to developmental robotics \cite{Lungarella03Developmental}. 
Developmental robotics is the interdisciplinary approach to the autonomous design of behavioural and cognitive capabilities in artificial agents that directly draws inspiration from developmental principles and mechanisms observed in children's natural cognitive systems \cite{Cangelosi18,Lungarella03Developmental}\footnote{Synonyms of Developmental Robotics include \textit{cognitive developmental robotics}, \textit{autonomous mental development} as well as \textit{epigenetic robotics}}. %

In this context, CL must consist of a process that learns cumulative skills and that can progressively improve the complexity and the diversity of tasks handled.

Autonomous agents in such settings learn in an open-ended \cite{Doncieux18} manner, but also in a continual way.  Crucial components of such developmental approach consist of learning the ability to autonomously generate goals and explore the environment, exploiting intrinsic motivation \cite{Oudeyer07} and computational models of curiosity \cite{Oudeyer18}.

We propose a framework to link continual learning to robotics. This framework also sets the opportunities for continual learning to have a framed mathematical formulation %
to present approaches in a clear and systematic way. %

First we present the context and the history of continual learning. Secondly, we aim at disentangling vocabulary around continual learning to have a clear basis. Thirdly, we introduce our framework as a standard way of presenting CL approaches to help transfer between different fields of continual learning, especially to robotics. 
Fourthly, we present a set of metrics that will help to better understand the quality and shortcomings of every family of approaches.
Finally, we present the specifics and opportunities of continual learning in robotics that make CL so crucial.

We kept the sections definitions, framework, strategies and evaluation general enough to both robotics and non-robotics domains. %
Nevertheless, the last section, \textit{Continual Learning for Robotics} \tim{(Section \ref{sec:robotics})} benefits from the content of previous sections to present the specificities of Continual Learning in the field of robotics.

\section{Definition of Continual Learning}

Given a potentially unlimited stream of data, a Continual Learning algorithm should learn from a sequence of partial experiences where all data is not available at once.
A non-continual learning setting would then be when the algorithm can have access to all data at once and can process it as desired.
Continual learning algorithms may have to deal with imbalanced or scarce data problems \cite{Sprechmann18}, catastrophic forgetting \cite{French99}, or data distribution shifts \cite{Gepperth16}.

\vin{As a more constrained version of on-line learning, CL needs to implicitly or explicitly account for knowledge fusion at different levels over time.}
Firstly, CL is required to support data-level fusion and, at the same time, be able to preserve learned knowledge from data that may disappear (e.g. due to inability to re-process certain data, due to the \textit{right to be forgotten} of EU GDPR\footnote{Art. 17 GDPR – Right to erasure (\textit{right to be forgotten}) \url{https://gdpr-info.eu › art-17-gdpr}}, or simply legacy reasons like for medical records). CL requires as well fusion at model level, since different tasks to be learned may require different model architecture components that in the end must act as one. %

\vin{Lastly, fusion needs also to occur at the knowledge or conceptual level, since memorization of raw data has to be avoided, but without incurring catastrophic forgetting.}
We consider continual learning a synonym of \textit{Incremental Learning} \cite{Gepperth16, Rebuffi16}, \textit{Lifelong Learning} \cite{Chen18, Thrun95} and \textit{Never Ending Learning} \cite{Carlson10, Mitchell15}. 
For the sake of simplicity, in the remaining of the article we refer to all Continuous, Incremental and Lifelong learning synonyms as Continual Learning (CL).

In this section we first present the history and motivation of continual learning, then we present several definitions of terms related to CL and, finally, we present challenges addressed by CL in machine learning.

\subsection{History and Motivation}

The concept of learning continually from experience has always been present in artificial intelligence and robotics since their birth \cite{turing09, weng01}. However, it is only at the end of the $20^{th}$ century that it has began to be explored more systematically. Within the machine learning community, the lifelong learning paradigm has been popularized around 1995 by \cite{Thrun95} and \cite{ring94}, while the robotics field only later catches up with a renewed interest in developmental robotics \cite{Lungarella03Developmental}.

Between the end of the 90s and the first decade of the $21^{st}$ century, sporadic attention has been devoted to the topic within the supervised, unsupervised and reinforcement learning domains. However, despite the first pioneering attempts and early speculations, research in this area has never been carried out extensively until the recent years \cite{Parisi18review, Chen18}. We argue that this is because there were more complex and fundamental problems to solve and a number of additional constraints:

\begin{itemize}
\item \emph{Lack of systemic approaches}: Machine learning research for the past 20 years has focused on statistical and algorithmic approaches on simple tasks (e.g., tasks where the distribution of data is assumed static). CL typically needs a systems approach that combines multiple components and learning algorithms in complex and dynamic tasks. The complexity of tasks and their multiple uses in continual learning greatly complicates training and evaluation procedures. Disentangling \emph{``static''} learning performance from continual learning side effects is important for the very incremental nature of the research and to facilitate comparison between approaches in this area.

\item \emph{Limited amount of data and computational power}: Digital data is a luxury of the $21^{st}$ century. Before the big data revolution, collecting and processing data was a daunting task. Moreover, the limited amount of computational power available at the time did not allow complex and expensive algorithmic solutions to run effectively, especially in a continual learning setting which undoubtedly makes learning more complex by having to deal with multiple tasks at the same time, as well as having to incorporate the concept of time into the learning process. 
\item \emph{Manually engineered features and ad-hoc solutions}: Before early 2000s and first works on representation learning, creating a machine learning system meant to handcraft features and finding ad-hoc solutions, which may differ significantly depending on the task or domain. Having a general algorithm with a more systematic approach seemed for a long time a very distant goal.
Manually engineered features is also a clear limitation to achieve autonomy, as new tasks need to have the same features or re-engineered ones.
\item \emph{Focus on supervised learning}: creating labeled data is probably the slowest and the most expensive step in most ML systems. This is why learning continuously has been for a long time not a viable and practical option.
\end{itemize}

The relaxation of these constraints, thanks to recent advancements and results in machine learning research, as well as the rapid technological progress witnessed in the last 20 years, have open the door for starting tackling more complex problems such as learning continually. %

We argue that the robotics community, which has always been intrigued by endowing embodied machines with lifelong and open-ended learning \cite{Doncieux18} of new skills and new knowledge, would highly benefit by the recent advances of ML in this area. Robotics applications in unconstrained environments, indeed, have always posed questions out of reach for previous machine learning techniques. On the other hand, CL developed in the context of robotics is involved in understanding the %
role and the impact of the concept of ``embodiment'' in intelligent machines that learn and think like humans. 

Learning, embodiment, and reasoning are presented as the three great families of challenges for robotics in \cite{Sunderhauf18}. We postulate that CL tackles the learning problem, taking into account the importance and constraints of embodiment. At best, CL would also benefit from reasoning in order to maximize the learning process. Thus, continual learning lies in the intersection of crucial robotics challenges.

\nat{Despite lifelong learning approaches existing in different ML disciplines (such as evolutionary algorithms for example \cite{Bellas09, Bellas10,Bredeche18,Bellas10cognitive}), in the rest of the article we focus on recent continual learning developments in the context of gradient-based neural network and deep learning approaches.}
For a more detailed description of many other classic approaches to continual learning with shallow architectures we refer the reader to \cite{Chen18}.

\subsection{Terminology Clarification}

In this section we aim at clarifying the distinction and similarities of continual learning with related topics and terms used in the literature.

\textbf{Online learning} 

Online learning is a special case of CL \cite{Kaeding16} where updates are done on per single data point basis and therefore, the batch size is one. 
\tim{Online learning algorithms are suited to scenarios where information should be processed instantly, either to adapt the model to learn as fast as possible or because data can not be saved.}

 

\textbf{Few-shot Learning}  

Few shot learning \cite{Lake11,Fei-Fei06} is the ability to learn to recognize new concepts based on only few samples of them. 
It may be used for continual learning problems when the number of data points is very low.
\tim{The extreme case of zero-shot learning 
consists of the ability to detect new 
classes 
while being trained with a disjoint set of classes \cite{Wang19}. 
} 

\textbf{Curriculum Learning} 

Curriculum learning \cite{Bengio09curriculum} is a training process that proposes a sequence of  more and more difficult tasks to a learning algorithm in order to make it able to learn, at last, a generally harder task. 
The sequence of tasks is designed in order to be able to learn the last one.
Both CL and curriculum learning learn on a sequence of tasks (or partial experience). However, in curriculum learning, tasks are chosen in a way that makes possible to learn tasks of different complexity, by taking into account the difficulty of them, while in CL, tasks are not voluntarily chosen nor ordered. 
\tim{Furthermore, while the interest of curriculum learning ultimately lies 
into solving the last task, the continual learning objective is to be able to solve all tasks.}

\textbf{Meta-learning}

Meta-learning \cite{Brazdi2008Metalearning} is a learning process that uses meta-data about past experiences, such as hyper-parameters, in order to improve its capacity to learn on new experiences.
\tim{It also learns several different tasks; however, its goal is not learning without forgetting but to progressively improve the learning efficiency while learning on more and more tasks. }
It is also called "learning to learn", and it can or not be used in a continual learning setting. 

\textbf{Transfer learning}

Transfer learning \cite{Pratt93,Finn17, Zhao17} is the ability to use what has been learned from a previous task on a new task. The difference with continual learning is that transfer learning is not concerned about keeping the ability to solve previous tasks. In computer vision, transferring what has been learned from a past environment to new environments would be often referred to as \textit{domain adaptation} \cite{Patel15,Csurka17}.

\textbf{Active Learning} 

Active learning is a special case of semi-supervised machine learning in which a learning algorithm is able to interactively query the user (or some other information source) to obtain the desired output labels for new data points \cite{Settles09, Burr10}. %
Active learning may be used in CL to query new examples and have control of the data the algorithm has access to.

\subsection{Challenges Addressed by CL} 

In this section we describe the specific problems addressed by continual learning; the kind of problems that arise when data cannot be assumed i.i.d., and when the hypothesis that the data distribution is static is not valid.

\subsubsection{Catastrophic Forgetting}

Catastrophic forgetting \cite{Mccloskey89,French99} refers to the phenomenon of a neural network experiencing performance degradation at previously learned concepts %
when trained sequentially on learning new %
concepts \cite{Mccloskey89}. Since by definition the continual learning setting deals with sequences of classes or tasks, the catastrophic forgetting is an important challenge to be tackled.  
Catastrophic forgetting might also be referred to as \textit{catastrophic interference}. The notion of interference is pertinent since the acquisition of new skills interferes with past skills by modifying important parameters.

\subsubsection{Handling Memories}

One of the main components that distinguishes two CL approaches is the way they handle memories. In order to deal with catastrophic forgetting, each strategy should find a way to remember what gradient descent will make forget.
Continual learning needs a mechanism to \textit{store} memories of past tasks, which can take very different forms. It is important to note that memories can be saved in different manners: as raw data, as representations, as model weights, regularization matrices, etc.
An efficient memory management strategy should only save important information, as well as be able to transfer knowledge and skills to future tasks.
In practice, it is almost impossible to know what will be important and what could be transferable in the future; a trade off should then be found between the precision of the information saved and the acceptable forgetting.
This trade-off problem is known as the stability/plasticity dilemma \cite{Mermillod13}.

An important challenge inherent to handling memories is to automatically assess them. Learning new tasks may lead to degradation of the memories. Furthermore, the memory process needs mechanisms to evaluate how the memories are degraded, i.e., how it forgets. As no more data and labels from past tasks may be available, this check-up might be very challenging.

\subsubsection{Detecting Distributional Shifts}

When the distribution is not stationary, a shift into the data stream is observed.
 When there is no external information concerning this shift, the CL model has to detect it, and account for fixing it by itself.
An undetected shift in the data distribution will irrevocably lead to forgetting.
Changes in the data distribution over time are commonly referred to as \textit{concept drift}. This idea is related to online change detection algorithms \cite{Sarkar98,Moens18} or Bayesian surprise \cite{Sun11} in ML. 
Two kinds of concept drift are defined \cite{Gepperth16}:
Virtual and real concept drift.
Virtual concept drift concerns the input distribution only, and can easily occur, e.g., due to imbalanced classes over time. 
Real concept drift, on the contrary, is caused by novelty on data or new classes, and can be detected by its effect, on e.g., classification accuracy.
However shift may also happen when the task change. In RL for example an agent may have to solve a new task. Then the shift is not exactly in the data distribution but in the supervision signal.
Regardless of where exactly the shift happened it has to be detected to avoid catastrophic interference with non related skills or knowledge.

\subsection{%
Learning Paradigms Orthogonal to Continual Learning}

In this section we describe the relationship of continual learning with respect to the main three, generally acknowledged machine learning paradigms: supervised, unsupervised and reinforcement learning.

\subsubsection{Supervised Continual Learning}
\label{subsubsec:sup}

Supervised learning is the machine learning problem of learning from input-output example pairs \cite{Russell09}. 
For each input-output pair ($X_t$,$Y_t$), the model should learn to predict $Y_t$ from $X_t$.
 $X_t$ is the input data, $Y_t$ is the supervision signal. Supervised continual learning is a particular case where the data is not available all at once. The function should then be learned from a sequence of data points in order to be able to map data to labels at the end of the sequence for the whole dataset. Supervised Continual Learning approaches have been mostly focused on classification \cite{Lopez-Paz17,Maltoni18,Kirkpatrick17}.
 
 While the study of continual learning in this context may help disentangling the complexity introduced by algorithms that learn continually, in the context of robotics, the lack of supervision does not allow, most of the time, to apply directly supervised methods.

\subsubsection{Unsupervised Continual Learning}
\label{subsubsec:unsup}

Unsupervised learning refers to machine learning algorithms that do not have labels or rewards to learn from.
In the context of robotics, unsupervised continual learning may play an important role in building increasingly robust multi-modal representations over time to be later fine-tuned with an external and very sparse feedback signal from the environment. 
In order to learn robust and adaptive representations with unsupervised learning, the main objective is to find suitable surrogate and meaningful learning signals, as robotics priors \cite{Jonschkowski14, Lesort19}, self-supervised models or curiosity driven techniques.

A particular unsupervised task learned in a continual learning setting is the generation of images. Image generation is achieved by training generative models to reproduce images from a dataset. In a CL setting, the distribution changes over time and the generative model should be able to produce at the end images from the whole distribution. This problem has been studied for various generative models (cf. section \ref{sec:CL_Strategies}) as adversarial models \cite{wu2018memory, lesort2018generative}, variational auto-encoders \cite{Nguyen17, Ramapuram17, achille2018life,Farquhar18, lesort2018generative} and standard auto-encoders \cite{Triki17,Zhou12}.

There is also a different relation between unsupervised learning and CL, since unsupervised models can be used to learn representations from vast amounts of data sources and can then generate such data (cf section \ref{subsub:gen_replay}). This capacity can then be used to perform CL for classification \cite{wu18incremental, Shin17, Triki17, lesort2018marginal} or reinforcement learning tasks \cite{caselles2018continual}. Another use case is using data generation as a data augmentation strategy.

\subsubsection{Continual Reinforcement Learning}
\label{subsubsec:reinforcement}

Reinforcement Learning is a machine learning paradigm where the goal is to train an agent to perform actions in a particular environment in order to maximize the expected cumulative reward. In traditional RL, the world is modeled as a stationary MDP: i.e., fixed dynamics and states that can recur infinitely often \cite{Ring05}\footnote{This MDP assumption was recognized and first removed in \cite{Ring05}}.
Since in general, complex RL environments have no access to all data gathered at once, RL could often be framed as a CL situation.  %
Moreover, RL borrows several tools used in CL models, such as approximating data to an i.i.d. distribution, via either \textit{i)} setting multiple agents or actors to learn in parallel \cite{Mankowitz18}, or \textit{ii)} using a replay buffer (or experience replay \cite{mnih15}), that is equivalent to a particular category of CL (rehearsal, see Section \ref{subsub:rehearsal}).
An analogy of a popular stable method in RL is PPO algorithm \cite{schulman17}, which constrains learning by using the Fisher information matrix to improve learning continually, in the same way as some CL strategies (e.g., EWC, see Section \ref{subsubsec:penalty}).
Most of Continual Learning approaches in RL have been applied in simulation settings such as Atari games \cite{Kirkpatrick17}. \nat{However, many 
approaches \cite{Traore19, Kalifou19, Bellas10, Bredeche18} also tackle use cases on real robots.}

\section{A Framework for Continual Learning}
\label{sec:framework}

Despite the rapidly growing interest in continual learning and mainly empirical developments of the recent years \cite{Parisi18review}, very little research and effort has been devoted to a common formalization of algorithms that learn continually in dynamic environments. However, the availability of a common ground for thoroughly evaluating and understanding continual learning algorithms is essential to reduce ambiguities, enhancing fair comparisons and ultimately better advancing research in this direction.

Being able to better compare and evaluate continual learning strategies, while still being general enough to overlook implementation-dependent details over different learning paradigms, becomes essential. This is specially true when targeting deployment of CL paradigms in real-word applications, such as robotics. Nowadays, %
despite the existence of a basic set of shared practices, many are the fundamental questions often overlooked in recent continual learning research. For example, questions about the data availability during training and evaluation, the amount of supervision with respect to the tasks separation and composition, as well as common but biased assumptions on the nature of the data among others. A list of questions of interest we would like to address and report are the following:

\begin{enumerate}[label=(\alph{*})]

    \item Data Availability

    \begin{itemize}[noitemsep]
        \item $\bm{Q_1}$: \tim{\emph{Does some data need to be stored? if yes, how and what for? (e.g. regularization, re-training, validation)?}}
        \item $\bm{Q_2}$: \tim{\emph{Is the algorithm tuned based on the final performance? I.e. is it possible to go back in time to improve performance?}}
        \item $\bm{Q_3}$: \emph{\tim{Are data distributions assumed i.i.d. at any point?}}
        \item{$\bm{Q_4}$: \emph{Is each task assumed to be encountered only once? }}
    \end{itemize}

    \item Prior Knowledge
    \begin{itemize}[noitemsep]
        \item $\bm{Q_5}$: \emph{Is the continual learning algorithm agnostic with respect to the structure of the training data stream? (e.g. number of classes, numbers of tasks, number of learning objectives...)}

        \item $\bm{Q_6}$: \emph{Does the approach need a pretrained model for the CL setting? If so, what is the new knowledge that needs to be acquired while learning continually?}
    \end{itemize}
    
    \item Memory and Computational Constraints
    \begin{itemize}[noitemsep]
        \item $\bm{Q_7}$: \emph{How much available memory does the algorithm require while learning? Does the memory capacity requirement changes as more tasks are learned?}
        \item $\bm{Q_{8}}$: \tim{\emph{Is the continual learning algorithm constrained in terms of computational overhead for each learning experience? Does the computational overhead increase over the task sequence? }}
        \item $\bm{Q_{9}}$: \emph{Is the continual learning algorithm agnostic with respect to the data type? (e.g. images, video, text,...)} 
        \item $\bm{Q_{10}}$: \emph{Is the continual learning algorithm able to handle situations where there is not enough time to learn?}
    \end{itemize}
    
    \item Amount/Type of Supervision
    
    \begin{itemize}[noitemsep]

        \item $\bm{Q_{11}}$: \emph{In the presence of multiple tasks, is the task label available to the algorithm during the training phase? And during evaluation?}
        \item $\bm{Q_{12}}$: \emph{Are all the data labeled? or only the first training set? Can the user provide sparse label/feedback (e.g. active learning) to correct the system errors?}
        
    \end{itemize}

    \item Performance Expectation
    
    \begin{itemize}[noitemsep]

        \item $\bm{Q_{13}}$: \emph{What is expected from the algorithm to remember at the end of the full stream? Is it acceptable to forget somehow, when task, context or supervision change?}
    \end{itemize}
    
\end{enumerate}

\tim{To summarize these questions, in any new CL algorithm proposition, it is fundamental to clearly describe the data stream, its use, the algorithm functioning, its assumed 
prior knowledge , and its requirements in terms of supervision, memory and computation.}

We will now propose a comprehensive and detailed framework to help distill and disentangle different approaches in different continual learning settings and help answer these questions.

Early theoretical attempts to formalize the CL paradigm are found in \cite{Ring05} as a combination between reinforcement learning and inductive transfer. More general framework approaches include the one on non i.i.d. tasks of \cite{Pentina15}. As in \cite{Pentina15}, we assume CL is tackling a \tim{probably approximately correct (PAC)} learnable problem in the approximation of a target hypothesis $h^*$ as well as learning from a sequence of non i.i.d. training sets. Our framework could also be seen as a generalization of the one proposed in \cite{Lopez-Paz17}, where learning happens continuously through a \textit{continuum} of data and a ``task supervised signal'' $t$ may be provided along with each training example.

In continual learning data can be conveniently seen as drawn from a sequence of distributions $D_i$, and thus the need to redefine a CL framework taking into account this important property is defined as follows.

\begin{mydef}
\textbf{Continual Distributions and Training Sets} 

\tim{In Continual Learning, $\mathcal{D}$ is a potentially infinite sequence of unknown distributions $\mathcal{D} = \{D_1, \dots, D_N\}$ over $X \times Y$, with $X$ and $Y$ input and output random variables, respectively. At time $i$ a training set $Tr_i$ containing one or more observations is provided by $D_i$ to the algorithm.}
\end{mydef}

As the framework hereby proposed is supposed to be general enough to cover the orthogonal and classical unsupervised, supervised and reinforcement learning approaches, $Tr_i$, as better detailed in Definition \ref{def:cla}, is a collection of training observations/data samples that act as signal of the joint distribution to be learned.

\begin{mydef}
\textbf{Task} 
\label{def:task}

A task is a learning experience characterized by a unique task label $t$ and its target function $g_{\hat{t}}^*(x) \equiv h^*(x,t=\hat{t})$, i.e., the objective of its learning.
\end{mydef}

\tim{It is important to note that the tasks are just an abstract representation of a learning experience represented by a task label. This label helps to split the full learning experience into smaller learning pieces. However, there is not necessarily a bijective correspondence between data distributions and tasks.}






\begin{mydef}
\label{def:cla}
\textbf{Continual Learning Algorithm}
Given $h^*$ as the general target function (i.e. our ideal prediction model), %
and a task label $t$, %
a continual learning algorithm $A^{CL}$ is an algorithm with the following signature: 
\begin{equation}
	\forall D_i \in \mathcal{D}, \hspace{20pt} A^{CL}_i:\ \ <h_{i-1}, Tr_i, M_{i-1}, t_i>  \rightarrow <h_i, M_i> 
\end{equation}

Where:
\begin{itemize}
	\item $h_i$ is the current hypothesis at timestep $i$, or, practically speaking, the parametric model learned continually.
	\item $M_i$ is an external memory where we can store previous training examples or partial computation not directly related to the parametrization of the model.
	\item $t_i$ is a task label, that can be used to disentangle tasks and customize the hypothesis parameters. For simplicity, we can assume $N$ as the number of tasks, one for each $Tr_i$.
    \item $Tr_i$ is the training set of examples. %
Each $Tr_i$ is composed of a number of examples $e_j^i$ with $j \in [1,\dots,m]$. Each example $e^{i}_j = <x^{i}_j, y^{i}_j>$, where $y^{i}$ is the feedback signal and can be the optimal hypothesis $h^*(x,  t)$ (i.e., exact label $y^{i}_j$ in supervised learning), or any real tensor (from which we can estimate $h^*(x, t)$, such as a reward $r^{i}_j$ in RL). 
\end{itemize}
\end{mydef}

It is worth pointing out that each $D_i$, can be considered as a stationary distribution. However, this framework setting allows to accommodate continual learning approaches where examples can also be assumed to be drawn non i.i.d. from each $D_i$ over $X \times Y$, as in \cite{Gepperth16,Hayes18NewMetrics}.

\begin{mydef}
\textbf{Continual Learning scenarios}
\label{def:scen} 
A CL scenario is a specific CL setting in which the sequence of $N$ task labels respects a certain ``task structure'' over time. Based on the proposed framework, we can define three different common scenarios:
\begin{itemize}
	\item \textsf{Single-Incremental-Task (SIT)}: $t_1 = t_2 = \dots = t_N $.
    \item \textsf{Multi-Task (MT)}: $ \forall i,j \in [1,.., n]^2, i\neq j \implies t_i \neq t_j$.
    \item \textsf{Multi-Incremental-Task (MIT)}: $\exists\ i,j,k:\ t_i = t_j$ and $t_j \neq t_k $.
\end{itemize}
\end{mydef}

Table \ref{tab:batch-examples} illustrates an example to clarify the definition of SIT, MT and MIT.

\tim{An example of Single-Incremental-Task (SIT) scenario is an ordinary classification task between cats and dogs, 
where the distribution changes through time. First, there may only be input images of white dogs and white cats, and later only black dogs and black cats. Therefore, while learning to distinguish black cats from black dogs the algorithm should not forget to differentiate white cats from white dogs. The task is always the same, but the concept drift might lead to forgetting.}


\tim{However, in a classification setting, a Multi-Task (MT) scenario would first consist of learning cats versus dogs, and later cars versus bikes, without forgetting. The task label changes when the classes change, and the algorithm can use this information to maximize its continual learning performance.}
\tim{The Multi-Incremental-Task (MIT) is the scenario where the same task can happen several times in the sequence of tasks, but such task is not the only existing one.}

\begin{table}[!htbp]
\centering
\caption{Example: Sequential task labels (corresponding to different distribution $D_i \in \mathcal{D}$) to reflect differences among CL categorization w.r.t. number and unicity of tasks for SIT, MT and MIT. 
Notice that a MIT setting requires breaking the constraint definition of SIT but also breaking the constraint definition of MT, i.e., it corresponds to the case where not all the tasks are considered having the same \textit{ID}, and not all the task are considered distinct.}
\label{tab:batch-examples}
\begin{tabular}{|l|c|c|c|} %
\hline
\textbf{Task ID/Session} %
&   \multicolumn{3}{|c|}{\textbf{CL settings}}  \\\hline
\textbf{Task ID} & \textbf{SIT} & \textbf{MT} & \textbf{MIT} \\\hline\hline
$t_1$ & 0 & 1 & 0   \\\hline  
$t_2$& 0 & 2 & 1   \\\hline 
$t_3$ & 0 & 3 & 0   \\\hline 
... & ... & ... & ...   \\\hline 
$t_i$ & 0 & i & ...   \\\hline 
\end{tabular}
\end{table}

In any learning problem (be it classification, RL or unsupervised learning), the ability to adapt to new concepts to be learned (from the \tim{PAC} ML framework \cite{Valiant84}),
as well as new instances of each concept, should be accounted. This is the objective of the next definition where we formally set three different settings an algorithm is required to manage, as they can have very high impact on the algorithm performance.


\begin{mydef}
\textbf{\tim{Task label and concept drift scenarios}}
\label{def:label} 
\tim{The task label can specify different assumptions made in a continual learning scenario.
We can define three main categories of task label assumptions regarding concept drift:}

\begin{itemize}
	\item \tim{\textsf{No task label}: Changes in the distribution are not signaled by any task label. The task is always the same (equivalent to SIT scenario)}.
    \item \tim{\textsf{
    Sparse task label}: Changes in the distribution are sparsely signaled by the task label. There are several tasks but changes in distribution may as well happen inside a task.}
    \item \tim{\textsf{Task label oracle}: Every change in the data distribution is signaled by the task label, which is given. } 
\end{itemize}
\end{mydef}

\begin{figure}[h]
    \centering
    \includegraphics[width=0.6\textwidth]{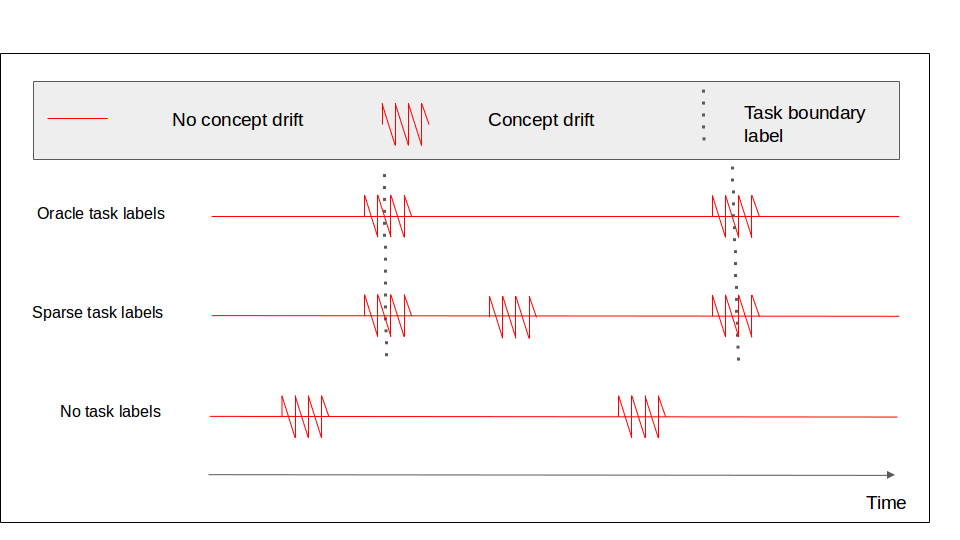}
    \caption{Task label and concept drift: illustration of the different scenarios.}
    \label{fig:concept_drift}
\end{figure}

\tim{We illustrate the different scenarios in Figure \ref{fig:concept_drift}.}

\begin{mydef}
\tim{\textbf{Availability of task label.} When a task label is provided, 
it is worth distinguishing among two different cases:}
\begin{itemize}
	\item \tim{\textsf{Learning labels}: Task labels are provided for learning only. At test time, inference should be done without knowing from which task a data point is coming from. }
    \item \tim{\textsf{Permanent labels}: The task labels are provided for learning, and it is assumed they will also be provided at test time for inference.}
\end{itemize}
\end{mydef}

\begin{mydef}
\label{def:cut}
\textbf{Content Update Type}
The nature of the data samples or observations contained in each $Tr_i$ can be conveniently framed in three different categories: 
\begin{itemize}
    \item \textsf{New Instances (NI)}: Data samples or observations contained in the training set at time-step $i$ 
    \nat{relate} to the same dependent variable \nat{$Y$} used in the past. %
	\item \textsf{New Concepts (NC)}: Data samples or observations contained in the training set at time-step $i$ 
    \nat{relate} to a new dependent variable \nat{$Y$} to be learned from the model. %
    \item \textsf{New Instances and New Concepts (NIC)}: Data samples or observations contained in the training set at time-step $i$ 
    \nat{relate} to both, already encountered dependent variables, and new ones \nat{($Y$)}. %
\end{itemize}
\end{mydef}

In order to exemplify the concept of \emph{Content Update Type} defined in Definition \ref{def:cut}, let us recover the aforementioned example of \tim{classification. If an algorithm learns the \textit{cat vs dogs} classification problem on a dataset and then new images of cat vs dogs are provided to the algorithm, we are then in a \textit{New Instances} case (NI), we have new data but no new concepts. If the new instances were of different classes (e.g. cars vs bikes) we then would face the \textit{New Concepts} case (NC). The new instances and new concepts case would then have been a mix of both new images of known and new classes.}

\nat{If a CL algorithm uses a network pretrained on a dataset, the features of such dataset will need to be accounted for as one more task or the same, depending on the distribution of new instances and new classes according to definitions \ref{def:scen} and \ref{def:cut}. In other words, using a pretrained model is similar to assume there is a task already learned by the model, and the new learning experiences of the algorithm are just a continuum of learning curricula. If there is any intersection between the pretraining and the new tasks, it should be reported in the setting description. The pretraining effect can then be estimated with the metrics proposed in Section \ref{sub:metrics}.}


\paragraph{Constraints}

\begin{constraint}
For every step in time, the number of current examples contained in the memory is lower than the total number of previously seen examples\footnote{I.e., if we could fit all previous examples in memory $M$, it would become a problem of scarce interest for the CL community, given that re-training the entire model $h_i$ from scratch would be always possible \cite{Kaeding16}}: 
    $\forall i \in [1,...,n], |M_i| \ll \biggl\lvert \bigcup\limits_{i=1}^{i-1} Tr_{i} \biggr\rvert$
\end{constraint}

\begin{constraint}
Memory and computation for each iteration step $i$ are bounded. Given two functions $ops()$ and $mem()$ that compute the number of operations and memory occupation required by $A_i^{CL}$, two reasonably small values \emph{max\_ops} and \emph{max\_mem} should exist, such that, for each $i$, $ops(A_i^{CL}) < max\_ops$ and $mem(h_{i−1} , M_{i−1}) < max\_mem$.
\end{constraint}

$max\_ops$ and $max\_mem$ are the max throughput, in number of operations, and the max memory capacity of the system running $A_i^{CL}$. Having a memory and computational bounds for each iteration $i$ is an important constraint for a continual learning algorithm. The reason is that the number of training sets $Tr_i$ can potentially be unlimited, and thus, computation and memory should not be proportional to the number of hypothesis updates $h_i$ over time. %
A finite upper bound should exist and be considered, especially with $n \rightarrow \infty$.

\paragraph{Relaxation and desiderata}

Given the difficult setting and the additional constraints imposed by Continual Learning with respect to the classic ``static'' setting, many researchers in the recent literature have proposed new CL strategies in slightly relaxed \cite{Rusu16progressive,Kirkpatrick17, Mallya18, Lopez-Paz17} yet reasonable settings: 

\begin{relaxation}
\textsf{Memory relaxation}: Removes the fixed memory bound constraint over $ops()$ and $mem()$. 
\end{relaxation}

\begin{relaxation}
\textsf{Computation relaxation}: Removes the fixed computational bound constraint $ops(h_i) < max\_ops$.
\end{relaxation}

In both cases we assume that for practical applications, a finite (and reasonable) number of tasks $N$ are encountered, hence, for many settings with a generous memory and computational bounds, many continual learning strategies that, in terms of complexity and memory usage, grow somehow proportional to the number of training sets $Tr_i$ may still be a viable option, especially if they can guarantee better performance. 

Having defined a formal framework for CL, we can therefore highlight a number of desiderata:

\begin{desideratum}
\textsf{Storage-Free Continual Learning}: Avoid the use of external memory $M$ to store raw data.
\end{desideratum}

\begin{desideratum}
\textsf{Online Continual Learning}:
Limit the size of each training set, moving towards online learning so that $|Tr_i| = 1$.
\end{desideratum}

Being able to learn without storing any raw data would mean a large step towards continual learning. In fact, getting rid of storing raw data means that the learning algorithm is able to extract information from the current task that may be not only useful and accurate for the actual task, but also transferable for the future.
 
In our biological counterparts, namely the brain, a system-level consolidation process is often thought to take place, where memories are encoded, stored and than retrieved for rehearsal purposes \cite{delvenne09}. However, the idea of storing high-dimensional perceptual data appears impractical given the incredible amount of information flowing into our brain every day from our multi-modal senses. Being able to process data online as well, is an important desideratum especially for reducing adaptation time and operational memory usage in an embedded or robotics setting.

\begin{desideratum}
\textsf{Task indicator free Continual Learning}: %
Learning continually without help of an external signal $t$ indicating the current task, in particular at test time, is strongly desirable.
\end{desideratum}

\section{Continual Learning Strategies}
\label{sec:CL_Strategies}

In this section we present a summary of the most popular continual learning strategies in the literature (see Fig. \ref{fig:Venn}). For a more in depth overview, we refer the reader to the recent overview in \cite{Parisi18review} that additionally exposes the bio-inspired aspects of existing continual approaches.

\subsection{Dynamic Architectures Approaches}
\label{subsec:archi}

The architecture of learning models has a strong influence on how they learn.
One approach to CL is to modify dynamically the architecture of a model to make it learn new concepts or skills without interfering with old ones. We present two types of dynamic architectures. Firstly, when the changes in the architecture are explicit; and secondly, when changes are implicit architectural changes by freezing weights. We also present an important architectural approach to CL: dual memory models.

\subsubsection{Explicit Architecture Modification}
\label{subsubsec:expl}

Explicit dynamics architecture gather all methods that add, clone or save parts of parameters of the models to avoid catastrophic forgetting.

Progressive neural networks \cite{Rusu16progressive} is one of the first approaches within this paradigm for deep neural networks. 
For each new task to be learned, a new model is created connected to all past ones. The goal of this new model is to learn the new task by using what was already learned by previous models, and so develop the new skills needed.
At test time, the proposed method needs to input data to all the neural networks previously created, and needs to know the task index to pick the right output.
Because the weights are used to connect neural networks together, the growth of parameters is quadratic w.r.t. the number of tasks. This growth is generally to be prevented. %
Instead, layers may be dynamically expanded in a single network without the need of re-training or freezing previously learned parameters, improving model capacity over time \cite{wang17}.

Another type of dynamic architecture strategy consists of  dynamically adding neurons for new tasks. As an example, output layers can be added in order not to change output parameters from previous tasks as in LWF approach \cite{Li17}. 
This method ensures that the output layer will not be modified; however, as the feature extraction layers are shared between tasks, some parameters risk to be modified and forgotten. In addition, at test time, the method needs the task label.

It is worth mentioning that we consider as \textit{dynamic architecture}, those approaches that adapt their architecture specifically with the aim of not forgetting, while similar mechanisms can be used for other purposes\footnote{If the architecture is changed without this objective, it is not considered as part of the CL approach. As an example, when new classes are available, we might choose to make the output size grow to handle these, without making it as a way to not forget.}.

\subsubsection{Implicit Architecture Modification}
\label{subsubsec:freeze}

Implicit architecture modification is the use of model adaptation for continual learning without modifying its architecture. This adaptation is typically achieved by inactivating some learning units or by changing the forward pass path. 

We categorize the fact of dynamically freezing weights as an implicit dynamic architecture approach. It is implicit because the architecture of the model does not change; however, the capacity of the model to learn new tasks does in an inevitable way.

Freezing weights consist of choosing some weights at the end of a task that will no more change in the future. The backward pass will not be able to tune them anymore; however, they can still be used in the forward pass. This method assures that these weights will not forget, and tries to keep enough free parameters to learn in the future \cite{Mallya17, Mallya18, Serra18}.  The difficulty lies in freezing enough weights to remember, but not too much to still be able to learn new skills.
The way weight freezing is implemented in PackNet \cite{Mallya17}, Piggypack \cite{Mallya18}  or HAT \cite{Serra18} is by defining a special mask for each task that is used to both protect weights when new tasks are learned, and to define which weights to use at inference time for a given task. 
The use of masks to freeze important weights can be referred to as hard attention process \cite{Serra18}. 
Weight freezing can also be used to keep the decision boundary of the output unchanged \cite{Jung16}. 

An alternative to a weight freezing when tasks change is to define a dynamics path inside the model in order to use a specific path for a specific task and not modify already learned weights. This is the idea exploited in \textit{PathNet} \cite{Fernando17}.

The use of implicit architecture modifications is not incompatible with explicit architecture modification as it is shown in \cite{Mallya18, Serra18}.

\subsubsection{Dual Architectures}

Dual approaches characterize architectures that are split in two models. One model is used in order to learn the actual task and should be easily adaptable, while the second model is used as a memory of past experiences. 
This approach can be linked to interactions between the hippocampus and neocortex to avoid catastrophic interference in mammals \cite{Mcclelland95}. The stable network plays the role of the neocortex, and the flexible one plays the role of hippocampus \cite{Furlanello16,Gepperth16, gepperth16bio, Maltoni18}. 

The use of dual architecture is explicit in many bio-inspired approaches such as \cite{Furlanello16, gepperth16bio, Parisi18, Sprechmann18,kemker18fearnet}. Dual architectures are extended in \cite{Sprechmann18} with the addition of an embedding model, and then, continual learning happens in the embedding space. The dual architecture can also be extended to more than two components, as in FearNet \cite{kemker18fearnet}, which takes inspiration from the basolateral amygdala from the brain to add a third component that is able to choose between the flexible and the stable memory for recall.

\subsection{Regularization Approaches}
\label{subsubsec:regu}

\subsubsection{Penalty Computing}
\label{subsubsec:penalty}

Regularization is a process of introducing additional information in order to prevent overfitting \cite{bhlmann2011statistics}. In the context of Continual Learning, the model should not overfit a new problem because it would make it forget it's previous skills.
The regularization approaches in continual learning consist in modifying the update of weights when learning in order to keep memory of previous knowledge.

Basic regularization techniques that could be used for CL are weight sparsification, dropout \cite{Goodfellow13}, and early stopping \cite{Maltoni18}. 
\tim{These simple regularization techniques reduce the chance of weights being modified, and thus decrease the probability of forgetting.}
More complex methods consist in searching for important weights inside the models and protect them afterwards to prevent forgetting.
The Fisher matrix can be used to estimate the importance of weights and produce an adapted regularization as for Elastic Weight Consolidation (EWC) approach \cite{Kirkpatrick17}. For efficiency purpose, EWC only use the diagonal of the Fisher matrix to estimate importance.  \cite{Ritter18} proposes an alternative to get a better estimation of the Fisher matrix using the Kronecker factorization.
EWC approach needs to have clear task delimitation to compute Fisher matrix at the end of the task, but  Synaptic Intelligence (SI) \cite{Zenke17} extended the method in an online learning fashion to relax this constraint.
\cite{Lee17} propose to use a regularization method called \textit{incremental moment matching} to overcome catastrophic forgetting. This method saves the moment posterior distribution of neural networks weights from past tasks and uses it to regularize learning of a new task. Two different declinations of this method are proposed: one with the use of first order moment \textit{IMM-mean} and one with second order moment \textit{IMM-mode}.

Another method to apply regularization for continual learning is the use of \textit{Conceptor} \cite{Jaeger14, He18}. Conceptor are memory mechanism that store learned patterns and representation. They are used to guide the gradient of the loss function to prevent forgetting and then favor modification for some weights and penalize others. 

The regularization methods have been shown to be efficient in reinforcement learning \cite{Kirkpatrick17}, classification \cite{Kirkpatrick17,Ritter18, Zenke17,He18} and also generative models \cite{Nguyen17, Seff17}. A limitation is that after several tasks the model may saturate because of a too high regularization, and finding a good trade-off between regularization that allows learning without forgetting may be hard.

\subsubsection{Knowledge Distillation}
\label{subsubsec:distillation}

Distillation techniques were introduced by \cite{Hinton15} in order to transfer knowledge from neural network A to neural network B. The idea is that after A has learned to solve a task, we want B to share this skill with A. We then forward the same input to both A and B and impose B to have the same output as A. Distillation should be more efficient than retraining B because A produces a soft-target that helps B to learn faster. In order to apply this method for continual learning, after network A learned to solve the first task, and while B is learning the second one, we distill knowledge from A to B. In the end, B should be able to solve both tasks. This and related methods have been used in various approaches \cite{wu2018memory,Schwarz18, Furlanello16, Rusu16distillation,Kalifou19,Traore19,Dhar19,michieli2019knowledge}. A drawback of distillation is that it generally needs to preserve a reservoir of persistent data learned for each task in order to apply distillation from a teacher model to a student model. %
Distillation can also be used to transfer policy learning from one model to another \cite{Rusu16distillation}.

\subsection{Rehearsal Approaches}
\label{subsub:rehearsal}

Rehearsal approaches gather all methods that save raw samples as memory of past tasks.

These samples are used to maintain knowledge about the past in the model. Ideally, those samples are carefully chosen in order to be representative of past tasks; by default, they can be randomly chosen.

The initial strategy is to save the representative samples and incorporate them in the new training set \cite{Rebuffi16, lesort2018generative}. In the second article samples are chosen randomly for continual learning of generative models but in \cite{Rebuffi16} the set is carefully sorted in order to keep the most representative samples into a coreset. This process allows to dynamically adapt the weights of the feature extractor and strengthen the network connections for memories already learned without forcing to keep previous weights. 

However, the coreset can also be used for regularization purpose and not just to be replayed from time to time along with new data in the learning process. 

For example, the coreset can be used for distillation  in \cite{Robins95} and in A-LTM (Active Long Term Memory Networks) \cite{Furlanello16} or to regularize the gradient when learning new tasks as in GEM (Gradient Episodic Memory)   \cite{Lopez-Paz17} and A-GEM (Averaged Gradient Episodic Memory) \cite{Chaudhry19}. \tim{Coresets have also been used to regularize the continual learning of a generative model in the CloGAN approach \cite{Rios19}.}
In a bayesian learning setting the coreset can be incorporated into the prior to regularize learning update as in \cite{Nguyen17}. The authors experimented the use of a coreset to create a variational continual learning model (VCL).

The disadvantage of rehearsal approaches is the utilization of a separate memory of raw and unprocessed data which is a vanilla way of saving knowledge that does not respect data privacy. Nevertheless it ensure that the memories are not degraded through time.

\subsection{Generative Replay}
\label{subsub:gen_replay}

Instead of modeling the past from few samples as it is done in \textit{Rehearsal} approaches, \textit{Generative Replay} approaches train generative models on the data distribution.
Therefore, they are able to afterwards sample data from past experience when learning on new data. By learning on actual data and artificially generated past data, they ensure that the knowledge and skills from the past is not forgotten.
These methods have also been associated with the term \textit{pseudo-rehearsal} \cite{Robins95} or \textit{Intrinsic Replay} \cite{Draelos17NeurogenesisDL}.
They could be understood as methods that perform \textit{regeneration} of samples or internal states, and thus, they can be associated with model-based learning, where the model learns the data distribution of past experiences.
The generative models is generally a GAN \cite{Goodfellow14Generative} as in \cite{wu18incremental, lesort2018generative, Shin17} or an auto-encoder as in \cite{Draelos17NeurogenesisDL, kemker18fearnet, caselles2018continual, Kamra17}.
 
A classical method %
implementing a generative replay normally makes use of dual models \cite{Kamra17, Shin17, wu18incremental, Farquhar18, kemker18fearnet}.
One frozen model generates samples from past experiences and another learns to generate and classify actual samples in addition to the regenerated ones. When a task is over, we replace the frozen model by the actual one, %
freeze it, and initialize a new model to learn next task.

Generative Replay models can be categorized into two different approaches: "Marginal Replay" and "Conditional Replay" \cite{lesort2018marginal}. Techniques using \textit{Marginal Replay} make use of standard generative models, while \textit{Conditional Replay} are a particular case of the former where the generative model is conditional. Conditional models can generate data from a specific condition, e.g. a class or a task. In continual learning, it allows then to choose from which past learning experience we want to generate. It is important for example to balance data in generated replay \cite{lesort2018marginal}.

While most of the Generative Replay based approaches are meant to solve classification tasks \cite{kemker18fearnet, Kamra17, Shin17, wu18incremental, Rios19}, some models use it for unsupervised learning \cite{lesort2018generative, wu2018memory} or reinforcement learning \cite{caselles2018continual}.

\subsection{Hybrid Approaches}
\label{subsec:hybrid}

Most CL approaches have an implicit dual architecture strategy, as they always have a slow learning and a fast learning mechanisms to learn continually.
For example, in rehearsal approaches (Section \ref{subsub:rehearsal}) the stable model role is played by a memory that stores samples, in generative replay approaches (Section \ref{subsub:gen_replay}) a generative model plays the role of stable model, in some regularization approaches (Section \ref{subsubsec:penalty}) the stable model is played by the Fisher matrix which saves important weights.

\begin{figure}[ht]
  \centering
  \includegraphics[width=0.6\textwidth]{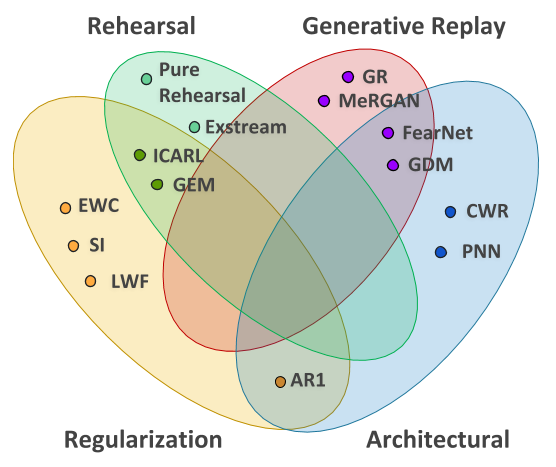}
  \caption{Venn diagram of some of the most popular CL strategies w.r.t the four approaches illustrated in Section \ref{sec:CL_Strategies}: CWR \cite{Lomonaco17}, PNN \cite{Rusu16progressive}, EWC \cite{Kirkpatrick17}, SI \cite{Zenke17}, LWF \cite{Li17}, ICARL \cite{Rebuffi16}, GEM \cite{Lopez-Paz17}, FearNet \cite{kemker18fearnet}, GDM \cite{Parisi18}, ExStream \cite{Hayes18MemoryEfficient}, Pure Rehearsal, GR \cite{Shin17}, MeRGAN \cite{wu2018memory} and AR1 \cite{Maltoni18}. Rehearsal and Generative Replay upper categories can be seen as a subset of replay strategies. Better viewed in color.}
  \label{fig:Venn}
\end{figure}

Moreover, most of continual learning approaches do not rely on a single strategy to tackle catastrophic forgetting.  As stated in previous sections, each approach offers advantages and disadvantages, but most of the times, combining strategies allows to find the best solutions.
We summarize in Table \ref{tab:classification} and Figure \ref{fig:Venn} the different approaches cited and the strategies they propose.

\newcolumntype{M}[1]{>{\centering\arraybackslash}m{#1}}

\begin{table}[!t]  %
\centering
\caption{Continual Learning  Main Strategies}
\label{tab:classification}
\begin{tabular}{|l |M{20mm} M{20mm} M{20mm} M{20mm}|} %
\hline
\multirow{2}*{\textbf{\footnotesize{References}}} &
\textbf{\footnotesize{Regularization}}&
\textbf{\footnotesize{Rehearsal}}  &
\textbf{\footnotesize{Architectural}}  &
\textbf{\footnotesize{Generative-Replay}} \\\hline\hline

\rowcolor{gray!20}
Zhou et al. \cite{Zhou12} & 
& %
& %
\checkmark & %
\\ %

Goodfellow et al. \cite{Goodfellow13} & 
  \checkmark &  %
& %
& %
\\ %

\rowcolor{gray!20}
Lyubova et al. \cite{lyubova15}  & 
 & %
\checkmark & %
& %
 \\ %

Rusu et al. \cite{Rusu16distillation}  & 
  \checkmark & %
& %
& %
 \\ %

\rowcolor{gray!20}
\tim{Camoriano et al.} \cite{Camoriano16} & 
\checkmark &%
\checkmark & %
& %
\\ %

Furlanello et al. \cite{Furlanello16} & 
  \checkmark & %
& %
  \checkmark & %
\\ %

\rowcolor{gray!20}
Li et al. \cite{Li17} (LwF)& 
  \checkmark & %
& %
  \checkmark & %
\\ %

Rusu et al. \cite{Rusu16progressive} (PNN) & 
& %
& %
  \checkmark & %
\\ %

\rowcolor{gray!20}
Jung et al. \cite{Jung16} & 
  \checkmark & %
& %
  \checkmark & %
\\ %

\tim{Aljundi et al.} \cite{Aljundi16}  & 
& %
& %
\checkmark & %
\\ %

\rowcolor{gray!20}
Rebuffi et al. \cite{Rebuffi16} (Icarl)  & 
  \checkmark & %
  \checkmark & %
& %
\\

Kirkpatrick et al. \cite{Kirkpatrick17} (EWC)  & 
  \checkmark  & %
& %
& %
\\ %

\rowcolor{gray!20}
Fernando et al. \cite{Fernando17}  & 
& %
& %
\checkmark & %
\\

Lee et al. \cite{Lee17}  & 
  \checkmark & %
& %
& %
\\ %

\rowcolor{gray!20}
Lee et al. \cite{Lee17Lifelong}  & 
  \checkmark & %
& %
& %
\\ %

Triki et al. \cite{Triki17}  & 
  \checkmark & %
& %
& %
\\ %

\rowcolor{gray!20}
Seff et al. \cite{Seff17}  & 
  \checkmark & %
& %
& %
\\ %

Shin \cite{Shin17} (DGR) & 
& %
& %
& %
  \checkmark \\ %

\rowcolor{gray!20}
Velez et al. \cite{Velez17}  & 
  \checkmark & %
& %
& %
\\ %

Lopez-Paz et al. \cite{Lopez-Paz17} (GEM)  & 
  \checkmark & %
  \checkmark & %
& %
\\ %

\rowcolor{gray!20}
Zenke et al. \cite{Zenke17} (SI) & 
  \checkmark & %
& %
& %
\\

Nguyen et al. \cite{Nguyen17} (VCL) & 
\checkmark & %
\checkmark & %
\checkmark  & %
\\ %

\rowcolor{gray!20}
Ramapuram et al. \cite{Ramapuram17}  & 
  \checkmark & %
& %
& %
  \checkmark \\ %

Mallya et al. \cite{Mallya17}  & 
& %
& %
  \checkmark & %
\\ %

\rowcolor{gray!20}
Kamra et al. \cite{Kamra17}  & 
& %
& %
& %
  \checkmark \\ %

Draelos et al. \cite{Draelos17NeurogenesisDL}  & 
& %
& %
& %
  \checkmark \\ %

\rowcolor{gray!20}
Serra et al. \cite{Serra18} & 
  \checkmark & %
& %
& %
\\ %

Mallya et al. \cite{Mallya18}  & 
& %
& %
  \checkmark & %
\\ %

\rowcolor{gray!20}
Parisi et al. \cite{Parisi18}  (GDM)& 
  \checkmark  & %
& %
  \checkmark & %
  \checkmark \\ %
 
He et al. \cite{He18}  & 
  \checkmark & %
& %
  \checkmark & %
 \\ %
 
\rowcolor{gray!20}
Hayes et al. \cite{Hayes18MemoryEfficient}  & 
 & %
\checkmark  & %
  & %
 \\ %
 
Wu et al. \cite{wu18incremental}  & 
& %
  \checkmark & %
& %
  \checkmark  \\

\rowcolor{gray!20}
Ritter et al. \cite{Ritter18}  & 
  \checkmark & %
& %
& %
\\

Schwarz et al. \cite{Schwarz18} &
& %
  \checkmark & %
& %
 \\ %

\rowcolor{gray!20}
Maltoni et al. \cite{Maltoni18} &
  \checkmark& %
 & %
  \checkmark& %
\\ %

Achille et al. \cite{achille2018life}  & 
& %
& %
  \checkmark & %
  \checkmark \\ %

\rowcolor{gray!20}
Wu et al. \cite{wu2018memory} (MeRGAN)  &   
  \checkmark & %
& %
& %
  \checkmark \\ %
  
\nat{Dhar et al.} & 
\checkmark & %
& %
& %
\\ %

\rowcolor{gray!20}
Lesort et al. \cite{lesort2018generative}  & 
& %
& %
& %
  \checkmark \\

Caselles-Dupr{\'e} et al. \cite{caselles2018continual}  & 
& %
& %
& %
  \checkmark \\ %

\rowcolor{gray!20}
\tim{Riemer et al.} \cite{riemer2018learning} (MER) & 
 \checkmark & %
  \checkmark   & %
 & %
  \\ %

\tim{Rios et al. }\cite{Rios19} (CloGAN) & 
\checkmark  & %
  \checkmark   & %
 & %
 \checkmark  \\ %

\rowcolor{gray!20}
Lesort et al. \cite{lesort2018marginal}  & 
& %
& %
& %
  \checkmark \\ %

Sprechmann et al. \cite{Sprechmann18}  & 
& %
  \checkmark  & %
  \checkmark  & %
 \\ %

\rowcolor{gray!20}
Kemker et al. \cite{kemker18fearnet} (FearNet) & 
& %
  & %
  \checkmark  & %
 \checkmark  \\ %
 
Chaudhry et al. \cite{Chaudhry19}  & 
  \checkmark & %
  \checkmark & %
& %
\\ %
 
\rowcolor{gray!20}
Kalifou1 et al. \cite{Kalifou19}  & 
  \checkmark & %
\checkmark  & %
& %
    \\\hline %

\end{tabular}
\end{table}

\section{Evaluation of Continual Learning Algorithms}
\label{sec:eval}
 
Before applying CL solutions to autonomous agents, they should be experimented and evaluated in simulation or toy examples. It is crucial to have a set of good evaluation metrics and benchmarks to assess if the approaches are scalable to real problems or may not solve harder ones. In this section we summarize existing evaluation methods and benchmarks and highlight some of them we believe worth using when targeting the deployment of practical CL applications.

\subsection{Evaluation Protocols and Benchmarks}%
\label{subsec:benchmarks}

\begin{table}[!t]

\caption{Benchmarks and environments for continual learning. For each resource, paper use cases in the NI, NC and NIC scenarios are reported.} 
\label{tab:benchmarks} 
\begin{tabular}{|p{50mm}|P{6mm}P{6mm}P{9mm}|P{50mm}|}
\hline
\textbf{Benchmark} & \textbf{NI}  & \textbf{NC} & \textbf{NIC} & \textbf{Use Cases} \\ %
\hline
\hline 

\rowcolor{gray!20}
Split MNIST/Fashion MNIST & 
 &  %
 \checkmark & 
 & %
\cite{lesort2018marginal, lesort2018generative, He18, Rios19} \\

Rotation MNIST & 
 \checkmark &  %
 & 
 & %
 \cite{Lopez-Paz17,lesort2018marginal, riemer2018learning}\\ %

\rowcolor{gray!20}
Permutation MNIST& 
 \checkmark &  %
 &
 & %
 \cite{Goodfellow13,Kirkpatrick17,Fernando17,Shin17,Zenke17,lesort2018marginal,He18, riemer2018learning}\\

iCIFAR10/100 & 
& %
\checkmark  & %
 & %
  \cite{Rebuffi16, Maltoni18,kemker18fearnet} \\

\rowcolor{gray!20}
SVHN & 
  & %
 \checkmark  &
 & %
 \cite{Kemker17, Seff17, Rios19}
 \\

CUB200 & 
\checkmark & %
 &
 & %
\cite{Lee17} \\

\rowcolor{gray!20}
CORe50 & 
 \checkmark & %
 \checkmark  &
 \checkmark  & %
\cite{Lomonaco17, Parisi18,Maltoni18} \\

iCubWorld28 & \checkmark & & & \cite{Pasquale15,lomonaco16} \\

\rowcolor{gray!20}
iCubWorld-Transformation  & & \checkmark & & \cite{Pasquale16,camoriano17a} \\

LSUN & 
& %
 \checkmark &
 & %
\cite{wu2018memory} \\

\rowcolor{gray!20}
ImageNet & 
& %
 \checkmark  &
 & %
 \cite{Rebuffi16, Mallya18}\\
 
 Omniglot & & \checkmark & & \cite{lake2015human,Schwarz18}\\

\rowcolor{gray!20}
Pascal VOC & & \checkmark & & \cite{michieli2019,Shmelkov17}\\


 Atari & \checkmark
& &
 & %
\cite{Rusu16distillation, Kirkpatrick17, Schwarz18}\\

\rowcolor{gray!20}
RNN CL benchmark & & & \checkmark  & \cite{Sodhani18}\\  %

CRLMaze (based on VizDoom) & \checkmark & & & \cite{lomonaco2019continual}\\
 DeepMind Lab &  \checkmark & & & \cite{Mankowitz18}\\

\hline

\end{tabular}
\end{table}
In continual learning, the difficulty of learning on a sequence of tasks is first of all dependant on the difficulty of each of the tasks separately. 
If a task is difficult to learn, a model will have to deeply modify its weights. If those weights contain knowledge from previous tasks, there is a high probability they will be degraded.
On the other hand, the risk of forgetting is also dependant on the likelihood of tasks occurring.
Indeed, after learning a task $T_t$, it is easier for a neural network to learn a radically different task $T_{t+1}$ without forgetting, than learning a task $T_{t+1}$ with similarities to $T_t$ \cite{Farquhar18}.

There are several kinds of similarities in a sequence of tasks:
\begin{itemize}
  \item Similarities in learning objectives: They occur when the objective is similar from task to task. For example, in a classification setting, when the same classes are used from one task to another (e.g. Permuted MNIST), or in RL, the same tasks need to be achieved in different environments.
  \item Similarities in features: the features from task to task are the same or very similar (e.g. Rotation MNIST).
\end{itemize}

Beyond the similarity among tasks and the learnability of each task, the availability of data is primordial to evaluate the difficulty of a benchmark. 
For convenience, most of the classical benchmarks assume that each task is available long enough to learn a satisfying solution.
Nevertheless, even when there is no constraint on the time to learn a task, data from the past can not be available again in the future. %
In several approaches, past data is used for model selection, however using the performance obtained on task $T_t$ to fine-tune a model that will learn on $T_0$ violates temporal causality \cite{pfulb2018a}. Data might be saved for later use as in rehearsal approaches, but this must be done before moving on to the next task.

Most CL benchmarks are benchmarks adapted from others fields, for instance:
\begin{itemize}
  \item \textbf{Classification}: MNIST \cite{LeCun10}, Fashion-MNIST \cite{Xiao2017}, CIFAR10/100 \cite{Krizhevsky09}, Street View House Numbers (SVHN) \cite{Netzer11}, CUB200 \cite{Welinder10}, LSUN \cite{Yu15}, ImageNet \cite{krizhevsky12}, Omniglot \cite{lake2015human} \nat{or Pascal VOC \cite{Everingham15} (object detection and segmentation)}.
  \item \textbf{Reinforcement Learning}: Arcade Learning Environment (ALE) \cite{Bellemare13} for Atari games, SURREAL \cite{Fan18} for robot manipulation and RoboTurk for robotic skill learning through imitation \cite{Mandlekar18}, \textit{CRLMaze} extension of VizDoom \cite{lomonaco2019continual} and DeepMind Lab \cite{Mankowitz18}. 
  \item \textbf{Generative models}: Datasets that prevail in this domain are the same as those used in classification tasks.

\end{itemize}

These datasets are then split, artificially modified (e.g., with image rotations or permutation of pixels) or concatenated together to create sequences of tasks and build a continual learning setting. As an example, permuted MNIST \cite{Kirkpatrick17} and rotated MNIST \cite{Lopez-Paz17} are continual learning datasets artificially created from MNIST.
\tim{Another possible continual learning scenario is the use of 
naturally non i.i.d. datasets (e.g. NICO \cite{He19}) or learning sequentially different datasets either on the same input space \cite{Lee17, Serra18} or in a multi-modal fashion \cite{Kemker17}.}
However, only few datasets, such as CORe50 \cite{Lomonaco17} or \cite{Sodhani18}, are specifically built with continual learning in mind.

In robotics, numerous datasets are often recorded in a online fashion through video. Therefore, they are suitable to evaluate continual learning algorithms. As an example, those proposed by \cite{Pasquale15,Pasquale16, azagra17} are composed of sequences of images captured during robotics object manipulation; they are used for classification and detection algorithms. 
A summary of the main datasets and examples of their applications can be found in Table  \ref{tab:benchmarks}.

\subsection{Continual Learning Metrics}
\label{sub:metrics}

Following the evaluation of an algorithm on a challenging benchmark, we should make sure that the  evaluation criteria are rigorous and cover the whole aspect of the full learning problematic. It is not enough to observe good final accuracy on an algorithm to know if it is transferable to a robotics settings. We should also evaluate how fast it learns and forgets, if the algorithm is able to transfer knowledge from one task to another, and if the algorithm is stable and efficient while learning. In this section we gather a set of metrics to rigorously evaluate a CL approach.

For a rigorous evaluation, we can assume to have access to series of test sets $Te_i$. The aim is to assess and disentangle the performance of our hypothesis $h_i$ as well as to evaluate if it is representative of the knowledge that should be learned by the corresponding training batch $Tr_i$.

For instance, one example of such evaluation is one of the first metrics proposed for CL \cite{Hayes18NewMetrics}; it consists of an overall performance $\Omega$ in a supervised classification setting. It is based on the relative performance of an incrementally trained algorithm with respect to an offline trained algorithm (which has access to all the data at once). In our notation, $\Omega$ is: %

\begin{equation}
\Omega = \frac{1}{N}\sum_{i=1}^{N}\frac{R_{i,i}}{R_{i,i}^{C}}.
\end{equation}  %

Where %
$N$ is the number of tasks encountered, $R^{C}_{i,j}$ is the potentially best accuracy we can have on $Te^C_i$ if the model was trained with all data at once, i.e. on $Tr^C_{i}$ (the accumulation of training sets $Tr^{C}_{t}$ from t=0 to t=i). $Te^C_{i}$ is the accumulation of all test sets $Te^{C}_{t}$ from $t=0$ to $t=i$.  %
 $\Omega$ = 1 indicates identical performance to an off-line cumulative setting; an $\Omega$ larger than one is possible when the offline model is worse than trained in a CL paradigm.

In \cite{Serra18}, instead, the authors try to directly model forgetting with the proposed \emph{forgetting ratio} metric $\rho$ after learning $i$ tasks, defined as:

\begin{equation}
\rho^{j\leq i} = \frac{1}{N} \sum_{i}^{N} \sum_{j}^{N} \left(\frac{R_{ij}-R_{j}^R}{R_{ij}^C -R_{j}^R}-1\right)
\end{equation}
Where, $R_{j}^R$ is %
the accuracy of a random stratified classifier using the class information of task $j$.

Always in the same sequential setting, in \cite{Lopez-Paz17} other three important metrics are proposed: \emph{Average Accuracy} (ACC), \emph{Backward Transfer} (BWT), and \emph{Forward Transfer} (FWT). In this case, after the model finishes learning about the training batch $Tr_i$, its performance is evaluated on all (even future) test batches $Te_j$.

The larger these metrics, the better the model. If two models have similar ACC, the preferred one is the one with larger BWT and FWT. Note that it is meaningless to discuss backward transfer for the first batch, or forward transfer for the last batch.
The metrics are extended for more fine grained, generic evaluation \cite{Diaz18} so that the original accuracy \cite{Lopez-Paz17} (as well as BWT and FWT) can account for performance at \emph{every timestep in time}. Accuracy is defined as:

\begin{equation}
A = \frac{\sum_{i=1}^{N}\sum_{j=1}^{i} R_{i,j}}{\frac{N(N+1)}{2}}
\end{equation}
where $R \in \mathds{R}^{N \times N}$ is the training-test accuracy matrix that contains in each entry $R_{i,j}$ the test classification accuracy of the model on task $t_j$ after observing the last sample from task $t_i$%
, Accuracy (A) considers the average accuracy for training set $Tr_i$ and test set $Te_j$ by considering the diagonal elements of  $R$, as well as all elements below it (i.e., averages $R_{i,j}$s where $i>=j$ see Table \ref{tab:acc-matrix-r}).
\begin{table}[!htbp]
\centering
\caption{Accuracy matrix $R$: elements accounted to compute A (white \& cyan), BWT (cyan), and FWT (gray). $R^{*} = R_{ii}$, \textbf{$Tr_i$} = training, \textbf{$Te_i$}= test tasks.}
\label{tab:acc-matrix-r}
\begin{tabular}{l|ccc} 
\textit{R}  & \textbf{$Te_1$} & \textbf{$Te_2$} & \textbf{$Te_3$} \\
\specialrule{.1em}{.05em}{.05em} 
\textbf{$Tr_1$} & $R_{1,1}$ & \cellcolor{Gray}$R_{1,2}$ & \cellcolor{Gray}$R_{1,3}$  \\ 
\textbf{$Tr_2$} & \cellcolor{LightCyan}$R_{2,1}$ &$R_{2,2}$ & \cellcolor{Gray}$R_{2,3}$  \\
\textbf{$Tr_3$} & \cellcolor{LightCyan}$R_{3,1}$ & \cellcolor{LightCyan}$R_{3,2}$ & $R_{3,3}$ \\
\specialrule{.1em}{.05em}{.05em} 
\end{tabular}
\end{table}
Backward Transfer (BWT) %
measures the influence that learning a task has on the performance on previous tasks. %
It is defined as the accuracy computed on $Te_i$ right after learning $Tr_i$ as well as at the end of the last task on the same test set (see %
 Table \ref{tab:acc-matrix-r} in light cyan).
\begin{equation}
BWT = \frac{\sum_{i = 2}^{N}\sum_{j = 1}^{i-1}(R_{i,j} - R_{j,j})}{\frac{N(N-1)}{2}}
\end{equation}
The original BWT \cite{Chaudhry18,Lopez-Paz17}
is extended %
into two terms to distinguish among two semantically different concepts (so that, as the rest of metrics, is to be maximized and in [0,1]). 
\begin{equation}
REM = 1- |min (BWT, 0)|
\end{equation}
i.e., \textit{Remembering}, and (the originally positive) BWT, %
i.e., improvement over time, \textit{Positive Backward Transfer}: %
\begin{equation}
BWT^{+} =  max (BWT, 0)
\end{equation}

Likewise, the FWT redefined to account for the dynamics of CL at each timestep is 
\begin{equation}
FWT = \frac{\sum_{i=1}^{j-1}\sum_{j=1}^{N} R_{i,j}}{\frac{N(N-1)}{2}}  %
\end{equation}

FWT accounts for the train-test accuracy entries $R_{i,j}$ above the principal diagonal of $R$, excluding it (see elements accounted in Table \ref{tab:acc-matrix-r} in light gray). Forward transfer can occur when the model is able to perform \textit{zero-shot} learning.

\nat{A Learning Curve Area (LCA) ($\in [0,1]$) metric to quantify the learning speed by a CL strategy is proposed in \cite{Chaudhry19}.} It uses the $b$-shot performance (where $b$ is the mini-batch number) after being trained for all the $N$ tasks: 

\begin{equation}
Z_{b} =  \frac{1}{N} \sum_{i=1}^{N}a_{i,b,i}  %
\end{equation}
where %
$a_{i,k,j} \in [0,1]$
is the accuracy evaluated on the test set of task $j$ after the model has been trained with the $k$-th mini-batch of task $i$. This amount is equivalent to previous accuracy matrix entry $R_{ij}$ but at a lower granularity of a batch level. $a_{i,k,j}$ is used to define a forgetting measure $\in [-1, 1]$ that quantifies the drop in accuracy on previous tasks \cite{Chaudhry18}. $f^k_j$ is the forgetting on task $j$ after the model is trained with all mini-batches up to task $k$:

\begin{equation} 
f_j^{k} = \max_{l \in {1,..,k-1}} a_{l, B_l, j} - a_{k, B_k, j}
\end{equation} 
where $B_i$ is all mini-batches corresponding to training dataset of task $k$ ($\mathcal{D}_k$).

$LCA_{\beta}$ is the area of the convergence curve $Z_b$ during training as a function of $b \in [0, \beta]$:
\begin{equation}
LCA_{\beta} =  \frac{1}{\beta + 1} \int_{0}^{\beta}Z_{b}db = \frac{1}{\beta +1}\sum_{b=0}^{\beta}Z_{b}
\end{equation}
The interpretation of LCA is intuitive: an $LCA_{0}$ is the average 0-shot performance (FWT), and $LCA_{\beta}$ is the area under the $Z_{b}$ curve, which is high if the 0-shot performance is good and if the learner learns quickly. LCA aims at disambiguating the performance of models that may have the same $Z_{b}$ or $A_T$, but very different $LCA_{\beta}$ because despite both eventually obtaining the same final accuracy, one may learn much faster than the other.

While forgetting and knowledge transfer could be quantified and evaluated in various ways, as argued in \cite{Farquhar18,Hayes18NewMetrics,Kemker17}, these may not suffice for a robust evaluation of CL strategies. For example, in order to better understand the different properties of each strategy in different conditions, especially for embedded systems and robotics, it would be interesting to keep track and unambiguously determine the amount of computation and memory resources exploited. In this context, the metrics proposed in \cite{Lopez-Paz17} are extended in \cite{Diaz18} to unify in a common evaluation framework different infrastructural and operational metrics. Other practical metrics included are Model Size (MS), Samples Storage Size (SSS) efficiency and Computational Efficiency (CE). \nat{We briefly describe them next.} 

The memory size of model $h_i$ is quantified in terms of parameters $\theta$ at each task $i$, $Mem(\theta_i)$; with the idea that it should not grow too rapidly with respect to the size of the model that learned the first task, $Mem(\theta_1)$:
\begin{equation}
MS = min(1, \frac{\sum_{i = 1}^{N}\frac{Mem(\theta_1)}{Mem(\theta_i)}}{N})
\end{equation}

Some CL approaches save training samples (or generative replay generated samples) 
as a replay strategy to not forget. The Samples Storage Size (SSS) efficiency establishes a metric for the memory occupation in bits by the samples storage memory $M$, $\mathit{Mem}(M)$, to be bound by the occupation of the total number of examples encountered at the end of last task: %
\begin{equation}
SSS = 1-min(1, \frac{\sum_{i = 1}^{N} \frac{\mathit{Mem}(M_i)}{\mathit{Mem}(D)}}{N})
\end{equation}
where $D$ is the lifetime dataset associated to %
all distributions $\mathcal{D}$.

A metric that bounds the Computational efficiency (CE) by the number of %
operations for training set $Tr_i$ is defined as: %
\begin{equation} %
CE = min(1, \frac{\sum_{i = 1}^{N} \frac{\mathit{Ops}\uparrow\downarrow(Tr_i) \cdot \varepsilon}{1+\mathit{Ops}(Tr_i)}}{N})
\end{equation}
where $Ops(Tr_i)$ is the number of (mul-adds) operations needed to learn $Tr_i$, $Ops\uparrow\downarrow$($Tr_i$) are the operations required to do one forward and one backward (backprop) pass on $Tr_i$, and $\varepsilon$ is a scaling factor (associated to the nr of epochs needed to learn $Tr_i$). %
Overall \textbf{$CL_{score}$} and \textbf{$CL_{stability}$} \nat{metrics are also finally proposed \cite{Diaz18}} in order to aggregate different criteria to be maximized that allow to rank CL strategies.
In order to assess a CL algorithm $A^{CL}$, each criterion to be optimized by the CL model, $c_i \in \mathcal{C}$ (where $c_i \in [0, 1]$) is assigned a weight $w_i \in [0,1]$ where $\sum_i^{\mathcal{C}}w_i=1$. Each $c_i$ 
\nat{is the average of $r$ runs, and the final \textbf{$CL_{score}$} to maximize is computed as:}
\begin{equation}
CL_{score} = %
\sum_{i=1}^{\#\mathcal{C}}w_i c_{i}   %
\end{equation}
\nat{where each final criterion $c_i$ is to be maximized by a CL algorithm.  
\textbf{$CL_{stability}$} is the pondered standard deviation of each CL metric \cite{Diaz18}:}
\begin{equation}
CL_{stability} = 1- \sum_{i=1}^{\#\mathcal{C}} w_i \sigma(c_{i})  %
\end{equation}

\tim{with $\sigma(c_{i})$ the standard deviation of criterion $c_{i}$.\\}
In future evaluation scenarios, particularly in robotics, stability is another important property that should be evaluated since in many robotic tasks and safety-critical conditions, potential abrupt performance drifts would be a major concern when learning continuously.
\nat{The metrics presented here can also be combined to assess higher-level capabilities. As an example, if we are to assess the \textit{scalability} of a CL algorithm, one could use a weighted average of \textit{SSS, MS}, and \textit{CE}.}

\nat{The metrics presented in a supervised classification context \cite{Diaz18}} can also be generalized with different performance measure $P$, instead of accuracy, and used in different settings such as reinforcement and unsupervised learning. For instance, they can be extended to RL; the underlying performance metric is, instead of accuracy, the accumulated reward on test episodes. In general in RL, cumulative reward plots over time are common norm to evaluate policy learning algorithms. Extra performance metrics in RL tasks will very much depend on the task being assessed, the reward function, and other evaluation metrics that act as evaluation \textit{proxies}, as it is common in semi/unsupervised learning settings. 

The evaluation of generative models in any setting is challenging.
Fréchet Inception Score (FID) \cite{Heusel17} is a common metric that compares features from generated data and true data.
Inception Score (IS) \cite{Salimans16} has also been widely used as a proxy to evaluate the quality of generative models. It measures if the class of generated samples are varied by making use of a model trained on ImageNet. One shortcoming of these scores is that they may be maximized by over-fitting generative models.
Another evaluation method is using generated data to train a classifier and evaluate its accuracy on a test set of true data \cite{Lesort18training}. The test accuracy, called Fitting Capacity (FC) gives a proxy on the quality of the generated data. 
Fitting Capacity and Fréchet Inception Score were used in a CL setting in \cite{lesort2018marginal, lesort2018generative}.\\
More methods for evaluating generative models are described and assessed more in depth in \cite{Borji18, Jiwoong18}; however, they have never been used in a CL setting.
In any case, the need for real data is mandatory in most evaluation schemes. In a CL setting, evaluating the generation of data from past tasks may need to violate the data availability assumption. The different metrics for generative models may then be useful tools for example for evaluating generative replay methods; however, they have to be manipulated carefully to be incorporated into the continual learning spirit.

\section{Continual Learning for Robotics}
\label{sec:robotics}

In the previous section we listed and described the different existing types of strategies to tackle continual learning. In this section, we will present real use cases of CL with an emphasis on robotics applications. First, we present why continual learning is crucial for robotics, and then, the challenges that robotics face in CL tasks. Finally, we present concrete robotic applications with potential insights to draw from CL. %

\subsection{Opportunities for Continual Learning in Robotics }
\label{subsec:opportunities}

A robot is an agent that interacts with the real world. It means that it can not go back in time to improve what it has learn in the past. %
These particularities of robotic platforms make them a natural playground for CL algorithms.
Furthermore, robots suffer from several constraints in terms of power or memory, and that is exactly what CL intends to optimize, in the way it addresses learning problems.
On the other hand, robots have rich information about their experiences. They are in control of their interaction with the environment, which may help them understanding the concept of causality, and extracting knowledge from different kinds of sensors (images, sound, depth...). This rich information helps machines to produce strong representations which are crucial for a well performing CL algorithm \cite{Lesort18}.

We could almost conclude that CL is born for robotics, and it may be true; however, today most of CL approaches are not robotics related and rather focus on experiments on image processing or simulated environments. Next section will present the challenges that make CL difficult to apply in robotic environments.

\subsection{Challenges of Continual Learning in Robotics} 
\label{subsec:challenges}

\subsubsection{Robotics Hardware}

The first challenge to deal with when doing any experiment with robots is the hardware.
Robots are known to be unstable and fragile. Robot failures are one of the main restrictions for researchers to propose new approaches on robotics tasks. They add unavoidable delay in any experiment and are expensive to fix.
Moreover, if the failure is not hardware but software, since it is not possible to reset the state of the robot automatically, manual help is often needed, e.g., to put back the robot in his starting position or recover it from an irrecoverable state.
Furthermore, most of the time building or buying a robot is itself quite costly.
Once the robot is correctly working, one new problem arises, which is its autonomy in terms of energy. This aspect is also a main difficulty to deal with when experiments need to be set. It is difficult to program long experiments without manually recharging the robot and making sure that it will not stop by a lack of power supply or failure.
Lastly, robots are embedded platforms and, consequently, have limited memory and computation resources, which should be carefully managed to avoid overflow.

The difficulties of using robots in experiments explain why there are so few approaches of continual learning with robots in the literature. In the next section, we will see how robotic environments challenge continual learning algorithms. 

\subsubsection{Data Sampling}
\label{subsec:sampling}

When a robot needs to learn a task in a known or unknown environment, it must collect its own training data in the real world \cite{Wong16}. Data serves as the basis for environment exploration and comprehension.
This problematic is exactly the same as the one met by RL algorithms \cite{Sutton98}. 
In infants, a crucial component of lifelong learning is the ability to autonomously generate goals and explore the environment driven by intrinsic motivation \cite{Oudeyer07, Cangelosi18}. 
Self-supervised approaches \cite{Pinto15, Levine16, Wong16, Shelhamer16} also help to automatically explore environments. 
Curiosity \cite{pathak18largescale} and self-supervision \cite{Doersch17} allow to search for new experiences (or data) and build a base of knowledge useful to achieve actual or future tasks via transfer learning \cite{Parisi18}.
As an example, manipulation tasks \cite{Kim19} such as grasping \cite{Pinto15}, reaching \cite{Raffin19, Colas18Curious}, pushing buttons \cite{Lesort19}, throwing \cite{Stulp14,Kim19} or stacking \cite{Colas18Curious} objects (cubes, balls...) are common complex tasks
built on comprehensive sets of experiments. %

Data gathered in this way can then be used on the fly in an online learning process or stored for later processing.

However, in order to improve learning algorithms the need for annotations or external help is crucial. In the next subsection we will describe the particular needs for annotations in robotics.

\subsubsection{Data Labeling}
\label{subsubsec:labeling}

As seen in previous section, gathering a varied set of raw data is already a difficult task. However, using it and understanding it is even more tedious. In this section, we detail different needs for labelling that autonomous agents such as robots need.
First of all, to understand its environment, a robot will need to apprehend the objects that compose it. To do so, the robot will need at some point that an external expert assesses that the object representation learned is good. This is the first kind of label the robot will need, i.e., object labelling \cite{Collet15, Craye18}.
Secondly, if we want the robot to perform a certain task, it will need to get information about the goals we gave it and also what it should not do. This is generally done by a reward function that defines credit assignment \cite{Minsky61}, 
or it can also be defined internally by more abstract rules such as self-supervision \cite{Gopnik01, Smith05}, intrinsic motivation or curiosity \cite{Oudeyer07, Schmidhuber10} as in \cite{Forestier2017intrinsically,Colas18, Craye18, Laversanne-Finot18}.
Thirdly, the robot should know when the task changes, and what task it should try to solve. This process consists of labelling the task; and the label is called the task identifier \cite{Lopez-Paz17}.

All these types of labels are not mandatory, but they drastically help and impact the learning process. 
The downside of labelling is that it is expensive and time consuming, which slows down the learning algorithms. To tackle those two problems, CL needs to find efficient solutions that can make the best out of the available labels for learning. 

The specific fields that aim at answering these questions are few-shot learning \cite{Lake11,Fei-Fei06} and active learning \cite{Burr10}. The former tries to grasp a concept from very few data points. Active learning aims at identifying and selecting the most needed labels in order to maximize learning. 
By combining optimization procedures in learning from few instances and minimizing the needs for labels, the field of robotics could be more suitable for leveraging continual learning settings in the real world. Furthermore, efficiency in learning reduces the risks of forgetting and degrading memories.

\subsubsection{Learning Algorithms Stability}
\label{subsubsec:learning_algo}

In continual learning, algorithms face several learning experiences in a row. From each learning experience, some memory should be saved to later prevent for not forgetting. The stability of learning algorithms is then crucial: if only one learning experience fails, the whole process may be corrupted. Moreover, if we respect the continual learning causality, we can not go back one or several tasks earlier in time in order to fix an actual problem. The corruption of one learning experience can lead to the corruption of memories and then to the model degradation when learning later tasks. The needs for robust mechanisms to validate or reject results of a learning algorithm is key to keep sane memories and weights; however, the instability of deep learning models must also be addressed to improve this drawback. As an example, generative models are powerful tools for continual learning but their instability may make them unsuitable for this kind of setting \cite{lesort2018generative}. Reinforcement learning algorithms are also known to be unstable and unpredictable, which is disastrous for continual learning. %

\subsection{Applications}
\label{subsec:applications}

\begin{figure}[ht]
  \centering
  \includegraphics[width=0.6\textwidth]{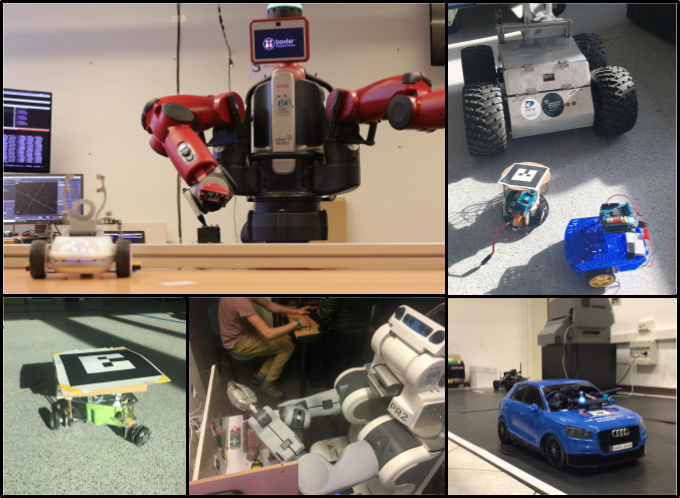}
  \caption{Sample tasks tested for unsupervised open-ended learning \cite{Raffin19,Doncieux18} and continual learning settings \cite{Kalifou19} in a couple of robotics labs, among others, from the DREAM project (\url{www.robotsthatdream.eu}).}
  \label{fig:robots}
\end{figure}

Real-word applications of continual learning are virtually unlimited. In fact, any learning algorithm that needs to deal with the real world will face a non i.i.d. data stream. This as well happens for autonomous robots that learn new manipulation tasks, for exploration policies, as well as for autonomous vehicles that need to learn and adapt to new circumstances %
\cite{Bojarski16,Codevilla17, Jaritz18,Rhinehart18}. Non-static settings are also faced by algorithms that learn how to predict trends based on data streams from internet user activities, e.g., among others, for advertisement or finance. This problem is likewise confronted when an already trained algorithms needs to acquire new knowledge without forgetting, e.g., recognize new classes for classification, anomaly detection, etc. However, in this section we focus on specific continual learning use cases on robotics.

\subsubsection{Perception}
\label{subsubsec:ap-classif}

While the world of perception is a multi-faceted topic at the very center of every application on autonomous sytems, the vast majority of CL algorithms in the literature are evaluated on object recognition tasks. Most models, indeed, are evaluated on datasets including static or moving objects. This is motivated by the fact that before any further action or policy, an autonomous agent (or robot) needs to identify the different component of its environment. In the case of classification, the robot may be pre-trained from an initial dataset. However, in any environment the robot would probably need to learn new objects from the new domain, and new variants (different poses, lighting, aspect) of already learned objects should be leveraged to improve its recognition \cite{lyubova15} capabilities. CL is crucial to tackle such dynamic scenarios. Initial progresses in this area have been proposed in \cite{Thrun95, Pasquale15, lomonaco16, camoriano17a, Lomonaco17}. \nat{Concrete Continual learning approaches to object segmentation can be found in \cite{michieli2019, michieli2019knowledge}, and in object detection in \cite{Shmelkov17}.}

Visual saliency for semantic segmentation and unsupervised object detection are other equally important applications in the context of perception which have been recently explored under continual learning and robotics settings \cite{Craye15}. RL-IAC (RL Intelligent Adaptive Curiosity), in particular, explores to learn visual saliency incrementally \cite{Craye18} with an articulated autonomous exploration technique based on curiosity to efficiently and continually learn a saliency model in a complex robotics environment tested in the real-world.

A classic problem in robotics within inherently continual learning settings are simultaneous localization and mapping (SLAM) \cite{cavallari2017fly} and navigation \cite{Thrun95}. In \cite{Thrun95}, using a HERO-2000 mobile robot with a radar sensor a continual learning algorithm based on explanation-based neural network learning (EBNN) is proposed to perform room mapping and navigation. Action models in EBNN \textit{explain} (in terms of previous experiences) and analyze observations to transfer task-independent (navigation) knowledge via predicting collisions and their prediction certainty. %

\subsubsection{Reinforcement Learning}
\label{subsubsec:ap-reinforcement}

In reinforcement learning the data distribution is dependent on the actions taken by the controlled agent. 
Therefore, since the actions taken are not random, data is not i.i.d. and the data distribution is not stationary.
In the context of reinforcement learning similar techniques to those proposed in CL are often adopted in order to learn over a data distribution which is approximately stationary. An example of such techniques is the use of a external memory for rehearsal purposes, also know as \textit{experience} or \textit{memory replay} buffer \cite{Lin92,Schaul15,Hayes18MemoryEfficient}.

The first challenge for RL is the extraction of representations to understand and compact what is relevant from the input data \cite{Lesort18}.
Continual learning of state representations for RL is intrinsically close to unsupervised learning or representation learning for classification; the methods used in both cases may then be very similar or worth leveraging across.%

The second RL challenge is learning a policy to solve a specific task. The CL challenge of policy learning is different because it often should take into account both state and context. Context is usually handled with recurrent neural networks, and this kind of model is not yet %
been studied extensively in CL; one example is in \cite{Sodhani18}. 
Different robot manipulation tasks such as grasping and reaching objects that are used as benchmarks can be seen in Fig. \ref{fig:robots} and, for instance, in state representation learning for robotics goal-based tasks \cite{Raffin19,Kalifou19}. 
These two challenges face continual learning problems, to learn representations and to learn policies from non stationary data distributions. However, it is worth distinguishing among both problems because learning and transfer between tasks are different challenges. Two tasks may need similar representations with different policies, while similar policies may require dissimilar representations.

In the context of robotics, fewer RL approaches have been proposed than in video-games or simulation settings. In particular, this is due to the low data efficiency of RL algorithms \cite{Raffin19}. 
We can still note several approaches that successfully tackle this problem, either in an end-to-end manner \cite{Kalashnikov18, Pinto15}, or by splitting the two challenges to address them separately, i.e., by first learning a state representation \cite{Lesort18} and later performing policy learning \cite{Finn15, Hoof16, Mattner12, Agrawal16, Duan17, Jonschkowski14}.
Nevertheless, a solution to this problem is to learn the policy in simulation and transfer it to deploy it in a real world robot \cite{Bojarski16, Rusu16sim2real, Gandhi17, Kalifou19}.

\subsubsection{Model-based Learning}
\label{subsubsap-ec:ers} 
Smoothly moving and interacting with always different, unpredictable environments, while constructing a coherent model of the external world, is one of the holy grail of robotics. For many years, researchers in this area have tried to propose robust and general enough sensory-motor solutions to complex problems such as navigation or object grasping. However, as it appears to be also true for humans, there will always be an environment or situation in which our biased model of the world fails and adaptation is needed.

Online (inverse dynamics) learning has also been applied in robotics, but generally not using deep learning. In \cite{Romeres16,Camoriano16}, the inverse and semiparametric dynamics of an iCub humanoid robot is learned in an incremental manner. This means both parametric modelling (based on rigid body dynamics equations) and nonparametric modelling (using incremental kernel methods) are used.
In \cite{Romeres18} it is shown that derivative-free models outperform numerical differentiation schemes in online settings when applied to non parametric (e.g. Gaussian processes with kernel function) model structures. %

In the pioneering work by \cite{Thrun95}, a model of both the external world and the robot itself is incrementally learned through reinforcement learning in complex navigation tasks on a real robot. However, incrementally and autonomously building a causal model of the external world, still remains a poorly explored topic in the context of robotics. Nevertheless, as it has been shown in recent RL literature, a model-based approach may be of fundamental importance in the real-world where millions of trials and errors are not always conceivable.

\section{Discussion and Conclusion}
\label{sec:discussion}

Several notions appear to be crucial to clearly describe learning algorithms in CL settings, fairly compare them and transfer them from simulation to real autonomous systems and robotics.
First of all, identifying the exact problem we want to solve, and what are the existing constraints is mandatory. The framework we introduce in Section \ref{sec:framework} should assist to achieve the characterization of these settings. This formal step helps finding the proper approach to use and identifying similarities with other settings.
Secondly, in the same spirit of defining what we want to learn, it is important to define the level of supervision we are able to give to our learning algorithm. For example, if we can give it the task label, or some kind of information about the structure of the input data stream (number of classes, type of data distribution, number of instances of each task, etc.). This point is also discussed in our proposed framework (Section  \ref{sec:framework}).
Finally, it is important to exactly clarify what is the expected performance of the algorithm. The set of metrics and benchmarks gathered in Section \ref{sec:eval} should help defining and articulating the dimension of evaluation for important properties worth considering in the development of embodied agents that learn continually.

For more concrete indications on what we consider worthwhile checking while creating a CL approach, we suggest a set of recommendations. \nat{After defining in Section \ref{sec:framework} a set of assumptions, constraints, relaxations and desiderata of CL algorithms, the following concrete measure and action-based guidelines aim at being taken into account as general advice} to palliate limiting factors of CL models in the literature.


\begin{recommendation} On-line capabilities: CL algorithms should not assume %
the number of total tasks to be solved is given beforehand. %
\end{recommendation}

\begin{recommendation} 
Learning complexity: We recommend keeping the learning model complexity below an upper bound of a linear growth in terms of the number of parameter growth when performing architectural dynamic changes. 
\end{recommendation}

\begin{recommendation} 
\tim{Scalability evaluation: In order to provide a proper evaluation of the scalability and continual learning performance, we recommend, as the authors from \cite{Farquhar18}, to evaluate algorithms on more than two tasks.}
\end{recommendation}

\begin{recommendation} %
Memory limitation: In order for realistic CL systems to be practical, they should not assume unlimited memory resources. 
\end{recommendation}

\begin{recommendation} Reporting metrics:  We recommend reporting as many metrics as possible and at least final performance, forward and backward (learning) transfer, the model's remembering capacity, model memory size, samples storage size, computational efficiency, CL score and stability metrics as described in Section \ref{sub:metrics}.
\end{recommendation}

 \begin{recommendation}
Offline baselines: we recommend the usage of publicly available baselines for metrics computation and fair assessment for reproducibility purposes.
 \end{recommendation}

\begin{recommendation} 
\tim{Ablation studies: we recommend reporting ablation studies to motivate as best as possible the different components and choices made in the CL algorithm such as initialization settings (using pre-trained network or not), optimization methods, hyper-parameters and surrogate losses used, etc.}
\end{recommendation}

\begin{recommendation}
Distributional shifts: We recommend to formally describe the mechanism to handle distributional shifts, not only when tasks change, but also among batches where data points conform to different distributions.
\end{recommendation}

\begin{recommendation} 
Benchmarks: We recommend the use of complex datasets with realistic and higher resolution scales than MNIST and CIFAR100; the use of the former is seen as a limiting factor and not a realistic robustness assessment method for CL (see Section \ref{subsec:benchmarks}). %
\end{recommendation}

\begin{recommendation} 
Report precisely and clearly how an approach learns and the assumptions it makes, as described in the framework (Section \ref{sec:framework}).
\end{recommendation}


To summarize, in this paper, we proposed a generalized framework to hold a variety of CL strategies and make easier the connection between machine learning and robotics in continual learning settings.
We reviewed the state of the art in continual learning and illustrated how to use the proposed framework to present various approaches. We also discussed benchmarks and evaluation techniques currently being used in continual learning algorithms.
We hope it helps the AI community to better categorize and compare approaches, as well as to smoothly adapt to today's industry problems. 
Machine learning and robotics are fields undergoing an aggressive development period. We believe that pushing them forward to find formalization solutions to facilitate transfer between both fields is critical in order to understand each other, and make them profit from each other's successes.

\subsubsection*{Acknowledgments}
This work is supported by the DREAM project\footnote{\url{http://www.robotsthatdream.eu}} through the European Union Horizon 2020 FET research and innovation program under grant agreement No 640891.

\bibliographystyle{abbrv} %
\bibliography{references}

\end{document}